\documentclass{article}

\usepackage[preprint]{neurips_2025}

\usepackage[utf8]{inputenc}
\usepackage[T1]{fontenc}
\usepackage{hyperref}
\usepackage{url}
\usepackage{booktabs}
\usepackage{amsfonts, amsmath, amssymb}
\usepackage{nicefrac}
\usepackage{microtype}
\usepackage{xcolor}         %
\usepackage{graphicx}
\usepackage{multirow}

\title{Unsupervised Learning of Density Estimates with Topological Optimization}

\author{Sunia Tanweer%
\\
Department of Mechanical Engineering \&\\Department of Computational Mathematics, Sciences and Engineering\\
Michigan State University\\
East Lansing, MI 48824\\
\texttt{tanweer1@msu.edu} \\
\And
Firas A. Khasawneh \\
Department of Computational Mathematics, Sciences and Engineering\\
Michigan State University\\
East Lansing, MI 48824\\
\texttt{khasawn3@msu.edu} \\
}

\begin{document}

\maketitle

\begin{abstract}
Kernel density estimation is a key component of a wide variety of algorithms in machine learning, Bayesian inference, stochastic dynamics and signal processing. However, the unsupervised density estimation technique requires tuning a crucial hyperparameter: the kernel bandwidth. The choice of bandwidth is critical as it controls the bias-variance trade-off by over- or under-smoothing the topological features. Topological data analysis provides methods to mathematically quantify topological characteristics, such as connected components, loops, voids et cetera, even in high dimensions where visualization of density estimates is impossible. In this paper, we propose an unsupervised learning approach using a topology-based loss function for the automated and unsupervised selection of the optimal bandwidth and benchmark it against classical techniques---demonstrating its potential across different dimensions.
\end{abstract}

\section{Introduction}
\label{intro}

Kernel Density Estimation (KDE) is a popular non-parametric method for estimating probability density functions given finite amount of data. It plays a crucial role in a wide range of applications, including classification~(\citet{Ghosh2006, Cho2020}), regression~(\citet{Hu2020, Chen2018}), stochastic dynamics~(\citet{Han2021, Kan2025, Kumar2016}), as well as neuroscience~(\citet{Mousavinejad2022, DtchetgniaDjeundam2013, StochMethods2009}). Despite the widespread usage, density estimation suffers from the fundamental challenge of tuning a hyperparameter---bandwidth---which controls the bias-variance tradeoff. An overly small bandwidth leads to an overfitted, spiky density function, whereas an excessively large bandwidth results in an underfitted estimate which fails to capture the underlying shape of the data. Consequently, optimal bandwidth selection remains an active area of research. Beyond global estimation accuracy, the bandwidth plays a decisive role in shaping the geometry and topology of the estimated density. Classical bandwidth selection methods---including plug-in estimators, cross-validation, and rules of thumb---typically optimize objectives such as the mean integrated squared error (MISE). While effective for global error control, these criteria are often misaligned with geometric or structural properties of the underlying density. For example, recent theoretical work by~\citet{Qiao2020} demonstrates that MISE-optimal bandwidths may fail to preserve important features in the level-set topology of the density. Even small perturbations in bandwidth can cause density level sets to merge or split, leading to abrupt changes in the topology of the estimated distribution. Thus, although MISE is statistically interpretable, it provides no mechanism for preserving the qualitative shape of the density.

Parallel research has established a strong connection between density estimation and topological data analysis (TDA)---highlighting that the topology of density level sets provides a meaningful and stable summary of the underlying distribution. Persistent homology applied to KDE level sets yields a robust multiscale representation of density structure and can even serve as a basis for hypothesis testing, density based clustering and topological uncertainty quantification~\citet{TDE}, with theoretical guarantees connecting persistence diagrams to underlying measure-theoretic properties. 

These observations expose a fundamental gap in the current KDE literature---although topological properties of density level sets are important in many
applications, no widely used bandwidth selection method is explicitly
topology-aware. Current methods either rely on restrictive assumptions
(e.g., local smoothness, Gaussianity), require model-specific tuning, or optimize criteria that do not reflect geometric or structural fidelity. As a result, the geometry and topology of the density estimate are largely uncontrolled.

Motivated by these deficiencies, we propose a fully unsupervised learning approach for the kernel bandwidth selection based on optimization of the topology features of the estimate. By leveraging persistent homology---the main algorithm in computational topology~(\citet{TamalDey2022})---which quantifies the shape of the data by capturing features such as connected components, loops, voids, and other higher-dimensional features, our method chooses the optimal bandwidth by minimizing a topology-based loss function. The inclusion of a quantifiable topology ensures that the density estimate is neither too complex nor too simplistic. We evaluate our method on  synthetic datasets across multiple dimensions, demonstrating its effectiveness in capturing underlying density structures. Unlike existing TDA-inspired methods~\citet{Fasy2018}, our approach requires no significance thresholds, no confidence interval estimation, and no bootstrapping. We evaluate the proposed method across a wide range of synthetic datasets spanning 1-4 dimensions, as well as on real image data from MNIST. Across these experiments, our estimator produces competitive results.

\section{Related Works}
\label{lit_review}

Bandwidth selection for kernel density estimation has been extensively studied and many approaches have been proposed, such as plug-in methods~(\citet{Delaigle2004, HALL1991}), cross-validation~(\citet{DUONG2005, Mugdadi2004}), bootstrapping~(\citet{Delaigle2004}), Bayesian and Monte Carlo approaches~(\citet{Cheng2018, Zhang2006}), adaptive techniques~(\citet{Abramson1982, Davies2017, Zmenk2023}), as well as neural-networks~(\citet{Wang2018}). However, due to the complexity and specificity of these techniques, the predominant approaches and software implementations in practice still largely remain the classical Scott's and Silverman’s rules of thumb~(\citet{scikitlearnKernelDensity}), which provide closed-form expressions for bandwidth estimation and computational efficiency. However, these rules assume approximately Gaussian structure and are known to perform poorly in multimodal or heterogeneous settings. Adaptive bandwidth methods alleviate these issues but introduce additional tuning parameters~\citet{Gao2022} and often require local density estimates, creating a circular dependency between the KDE and its bandwidth. Another fundamental limitation shared by nearly all methods is the focus of MSE or likelihood, which emphasizes global statistical accuracy but does not explicitly preserve the geometric or topological structure of the distribution. As noted earlier in~\citet{Qiao2020}, even MISE-optimal bandwidths can produce density level sets with incorrect topology.

Topological data analysis has emerged as a promising tool for describing the shape of data~(\cite{TamalDey2022}), and has recently been shown to be able to reliably quantify the topology of probability distributions even in high-dimensions~(\citet{Tanweer2024})---surpassing the need to visualize them. Our proposed method builds on these advancements by integrating topological features of the density estimate in the bandwidth selection process with openly available computational tools in topology. By leveraging persistent homology, we define a loss function minimizing which provides a topology-informed bandwidth for density estimation by maximizing influence of the significant topological features on the loss function. The closest precursor to our work was proposed in~(\citet{Fasy2018}), however their proposed loss function is not regularized and requires setting a significance level for confidence interval estimation with bootstrapping. Our method neither requires fixing any significance level, tuning any hyperparameters nor any bootstrapping of data, and bridges the gap by introducing a topology-informed loss function. 

\section{Proposed Method}
\label{method}

In this section, we introduce our approach for selecting the optimal bandwidth in KDE using computational topology. Unlike traditional methods that rely on heuristics or simplifying assumptions, our method formulates bandwidth selection as an optimization problem based on the topology of the estimated density function.

\subsection{Mathematical Preliminaries}

\subsubsection{Cubical Complexes}
Cubical complexes are constructed from elementary intervals---either unit intervals $[u, u+1]$ or degenerate intervals $[u, u]$ where $u \in \mathbb{Z}$. An $n$-dimensional cube is defined as $I = \prod\limits_{i = 1}^{K}[u_i, u'_i]$ where $u'_i \in {u_i, u_{i+1}}$. A cube's dimension equals its non-degenerate intervals. When cube $\sigma$ is a subset of $\tau$, we say $\sigma$ is a face of $\tau$ ($\sigma \leq \tau$). A cubical complex is a collection where including a cube implies including all its faces.

\subsubsection{Homology}

Homology quantifies topological features of spaces by analyzing structures like connected components, loops, etc. For a topological space $X$ composed of simplices or cubes, $p$-dimensional homology $H_p(X)$ examines how these building blocks interconnect across dimensions. The dimension of each homology group is called the $p$-th Betti number, denoted $\beta_p$, which quantifies the number of $p$-dimensional features in the space. $H_0$ represents connected components, $H_1$ represents loops, $H_2$ represents voids and higher-dimensional topological features can exist as $H_n$.

The $p$-th chain group $C_p(X)$ represents collections of $p$-dimensional elements, while the boundary map $\partial_p$ defines their connectivity relationships. Homology $H_p(X)$ is defined as $\text{Ker}(\partial_p) / \text{Im}(\partial_{p+1})$, where $\text{Ker}(\partial_p)$ contains cycles (closed structures) and $\text{Im}(\partial_{p+1})$ contains boundaries. This quotient identifies distinct topological features at dimension $p$.

\subsubsection{Persistent Homology}
For a function $f \to \mathbb{R}$ on a topological space, we analyze how homology changes across parameter values.
In a super-level filtration $X^{a_n} \subseteq X^{a_{n-1}} \subseteq \cdots \subseteq X^{a_1}$ where $X^a = f^{-1}[a,\infty)$ induces $H_p(X^{a_n}) \to H_p(X^{a_{n-1}}) \to \cdots \to H_p(X^{a_1})$.
These filtrations decompose into pairs $(b,d)$ representing birth and death of each topological feature, where $b$ is when the feature appeared and $d$ when it disappeared. The persistence (aka life) of each point is defined as $l = d - b$. The set of all such pairs, $D = \{(x, y): x = b, y = d, x>y\}$, is called the Persistence Diagram (PD). Higher life points are further away from the diagonal of the PD and correspond to a greater significance of the topological feature. 

\subsubsection{Persistence of Cubical Complexes}
\begin{figure}[!htbp]
  \centering
  \includegraphics[width=0.3\linewidth]{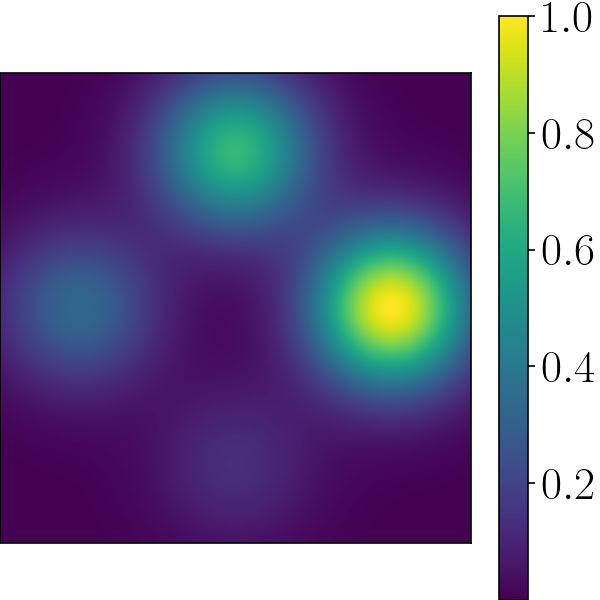}
  \includegraphics[width=0.27\linewidth]{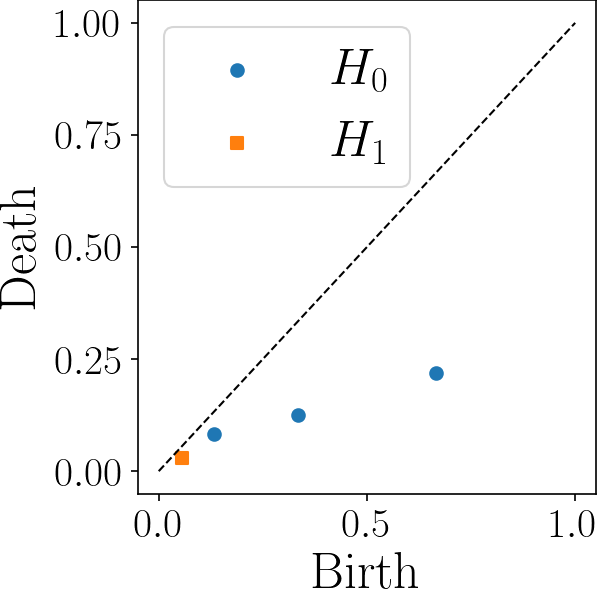}
  \includegraphics[width=0.27\linewidth]{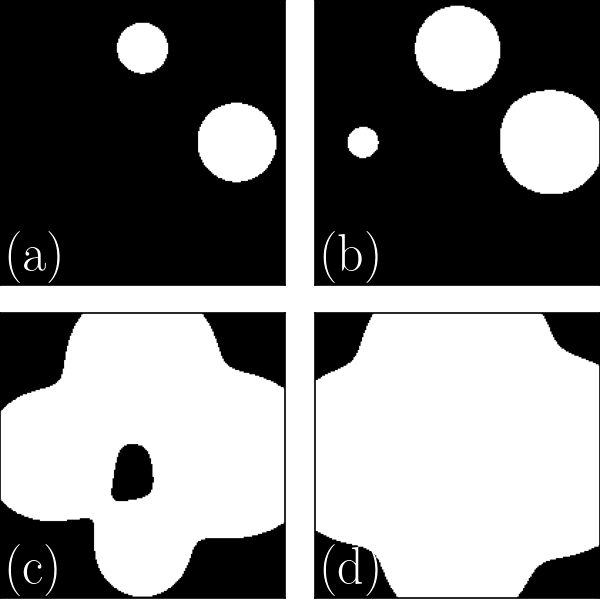}
  \caption{Super-level cubical persistence of (unit-normalized) quadmodal gaussian distribution. Figures correspond to (a) $K^{0.5}$ (two connected components), (b) $K^{0.3}$ (three connected components), (c) $K^{0.05}$ (one connected component and one loop) and (d) $K^{0.01}$ (one connected component).}
  \label{fig:cubical_example}
\end{figure}

For an $m \times n$ matrix $M$ (e.g., grayscale image), we define its domain as cubical set $K=\mathcal{K}([0,m]\times[0,n])$ with function $f \to \mathbb{R}$ where $f(s_{i,j})=M_{i,j}$ for cube $s_{i,j}=[i,i+1]\times[j,j+1]$. For lower-dimensional elements $P$, $f(P)=\max_{s,j>P} M(s_{i,j})$. The super-level set $K^a = f^{-1}[a, \infty)$ forms a cubical complex, enabling computation of super-level persistence as $a$ varies. See Fig.~\ref{fig:cubical_example} for an example. 

\subsection{Kernel Density Estimation}

Given a dataset $X = \{x_1, x_2, \cdots, x_n \} \subset \mathbb{R}^d,$ the kernel density estimate $\hat{f}_h(x)$ is computed using a kernel with bandwidth 
$h$
$$\hat{f}_h(x) = \frac{1}{nh^d} \sum_{i = 1}^n K (\frac{x-x_i}{h}),$$
where $K(\cdot)$ is the kernel. Throughout this work, the kernel is assumed to be Gaussian, that is, $K(x) = \frac{1}{\sqrt{2\pi}}\exp{(-\frac{x^2}{2})}$---although our method does not require or assume it. 

\subsection{Loss Function}

The central idea of our approach is that the kernel bandwidth should be chosen so that the KDE preserves the shape of the underlying density. The choice of $h$ significantly impacts the estimated density, affecting the smoothness and accuracy of the estimate. To quantify the effects of bandwidth selection, we compute the superlevel cubical PD of $\hat{f}_h(x)$ and define a loss function $\mathcal{L}(h)$ which balances two competing properties of the PD: simplicity and complexity. A very smooth density estimate results in a PD with only a few features (often a single dominant point with high persistence), indicating excessive smoothing---we quantify this using the Betti numbers of the PD. While an overly complex density estimate results in many short-lived topological features, indicating overfitting---we measure this by summing the lifetimes of all features in the PD (also called the total persistence). However, Betti number $\beta$, due to its discrete nature, is non-differentiable and does not allow for automatic differentiation. Therefore, we replace it with a smooth surrogate to compute the feature count accoring to

\[
\mathrm{count}(h)
=
\sum_{i = 1}^n
\sigma\!(\ell_i),
\]
where $\sigma(x) = \frac{1}{1+\exp{(-x)}}$ is the sigmoid function.

This leads to the loss function
\[
\mathcal{L}(h) = \alpha_{\mathrm{count}}\mathrm{count}(h) - \alpha_{\mathrm{TP}}\mathrm{TP}(h),
\]
where $\mathrm{TP}(h) = \sum_{i = 1}^n l_i$. This loss balances both the simplicity and complexity of the KDE. Unit-normalizing the density function before computing cubical persistence ensures that both $\mathrm{TP}(h)$ and $\mathrm{count}(h)$ have the same scale---allowing us to set $\alpha_{\mathrm{count}} = \alpha_{\mathrm{TP}} = 1$.

Hyperparameter sensitivity study (see Appendix~\ref{sec:sensitivity}) shows the results to be stable with respect to the three hyperparameters $\alpha_{\mathrm{count}}$ and $\alpha_{\mathrm{TP}}$ as well as grid size, while an ablation study (see Appendix~\ref{sec:ablation}) provides evidence of the importance of all terms in the loss. Hence, for all experiments, we use the loss function
\[
\centering
\boxed{
    \mathcal{L}(h) = \mathrm{count}(h) - \mathrm{TP}(h).
}
\]

\subsection{Optimization}

Persistent homology is stable under small perturbations of the underlying function~\citet{CohenSteiner2006}, and as a consequence the superlevel-set topology of $\hat{f}_h$ often remains unchanged across a wide interval of bandwidth values. When $h$ varies within such a region, the persistence diagram changes only minimally, causing both the feature count and total persistence to exhibit \emph{plateaus} rather than sharp minima. In practice, this means that $\mathcal{L}(h)$ may attain nearly identical values for a continuum of bandwidths that all induce the same topological structure. Without additional guidance, the optimizer is free to drift anywhere within this topologically equivalent plateau. To help with this, we use auto-differentiation with a Stochastic Gradient Descent (SGD) optimizer. 

\section{Experiments}
\label{sec:results}

We conducted experiments\footnote{The experiments were conducted on a system equipped with dual Intel Xeon Platinum 8260 processors, each running at a base clock speed of 2.40 GHz. The system has a total of 48 CPU cores---each having its own 1.5 MB L1 cache, 1.5 MB L2 cache, and the system as a whole has 71.5 MB of shared L3 cache.} on synthetic data sampled from known distributions, and benchmarked against a broad collection of commonly used bandwidth-selection techniques in statistics and machine-learning. These included Scott’s rule (the multivariate normal-reference bandwidth)~\citet{Scott2015}, Silverman’s rule of thumb~\citet{Silverman2018} and its multivariate normal-reference extension (NRR)~\citet{Scott2015}, as well as several classical cross-validation approaches such as maximum-likelihood cross-validation (ML-CV)~\citet{Duin1976}, least-squares cross-validation (LSCV)~\citet{Jones1996}, and biased cross-validation (BCV)~\citet{Jones1996}. We further compared against the Improved Sheather-Jones (ISJ) plug-in estimator (applicable only in one dimension)~\citet{Sheather2004}, Botev’s FFT-based projection estimator (BotevProj)~\citet{Botev}, and the diagonal plug-in bandwidth selector (PluginDiag)~\citet{Jones1996}. Together these represent the standard repertoire of reference-rule, plug-in, and cross-validation bandwidth estimators, providing a comprehensive baseline for evaluating our topology-driven approach. The KDEs generated by each are compared against the true probability density using the Earth Mover's Distance (EMD) and the Kullback-Leibler Divergence (KLD). Both EMD and KLD offer distinct perspectives on the dissimilarity between the estimated probability distributions. EMD quantifies the minimum amount of ``work" required to transform one distribution into the other. In contrast, KLD measures the information loss when one distribution is used to approximate another. While EMD provides a more geometric measure of dissimilarity, KLD offers an information-theoretic view, making them complementary tools. 

\subsection{One-Dimensional Datasets}

In 1D, we tested two distributions: a bimodal gaussian and a mixture of gaussians and cauchy distributions (Eqs.~\ref{eq:1d_1}-\ref{eq:1d_2}). For both, 500 simulations were run with 5000 points sampled in each, with KDE computed on a grid size of 200. In the former case, the TDA bandwidth achieves the lowest KLD, and the classical LSCV/BCV kernels produce the smallest EMD values, with TDA close behind. For the latter distribution, all plug-in and rule-of-thumb methods over-smooth. TDA does not match ISJ, which is known to excel on 1D problems, nor ML-CV in some runs, but it consistently provides a good compromise between fidelity and smoothness. Importantly, TDA shows low variance compared to ML-CV and retains stability across trials, while outperforming all classical rules (Scott, Silverman, Plugin, Botev). See Table~\ref{tab:1D} for a quantitative comparison of all methods.

\begin{figure}[!htbp]
\centering
\includegraphics[width=1\linewidth]{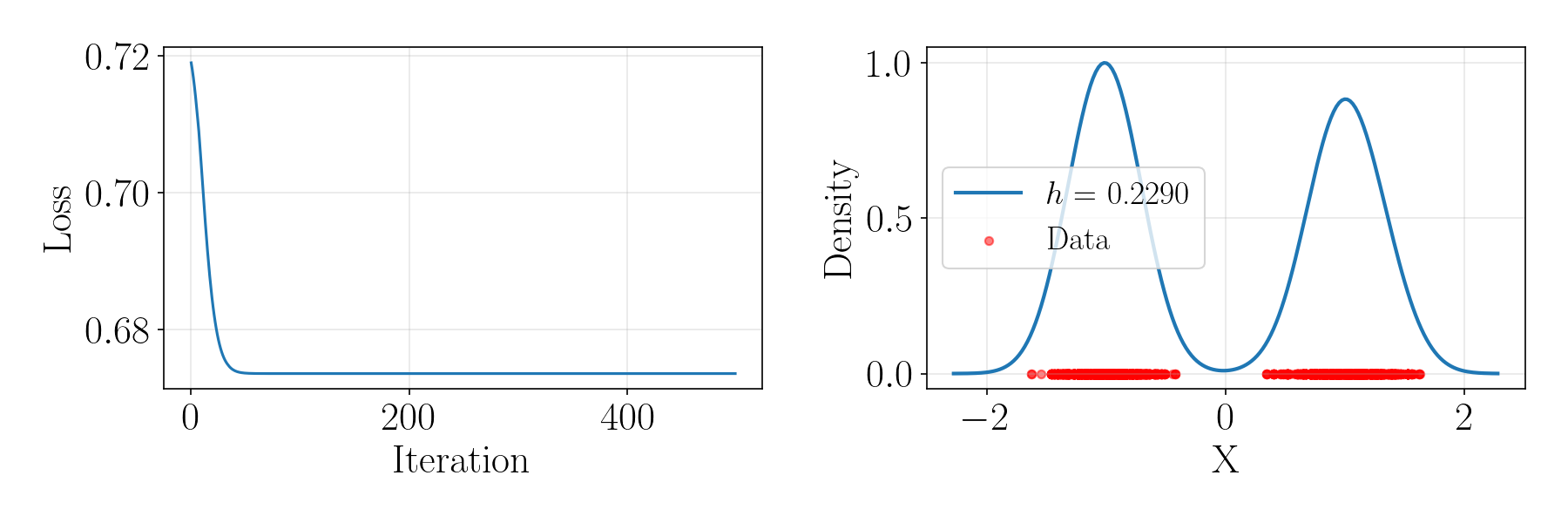}
\includegraphics[width=0.9\linewidth]{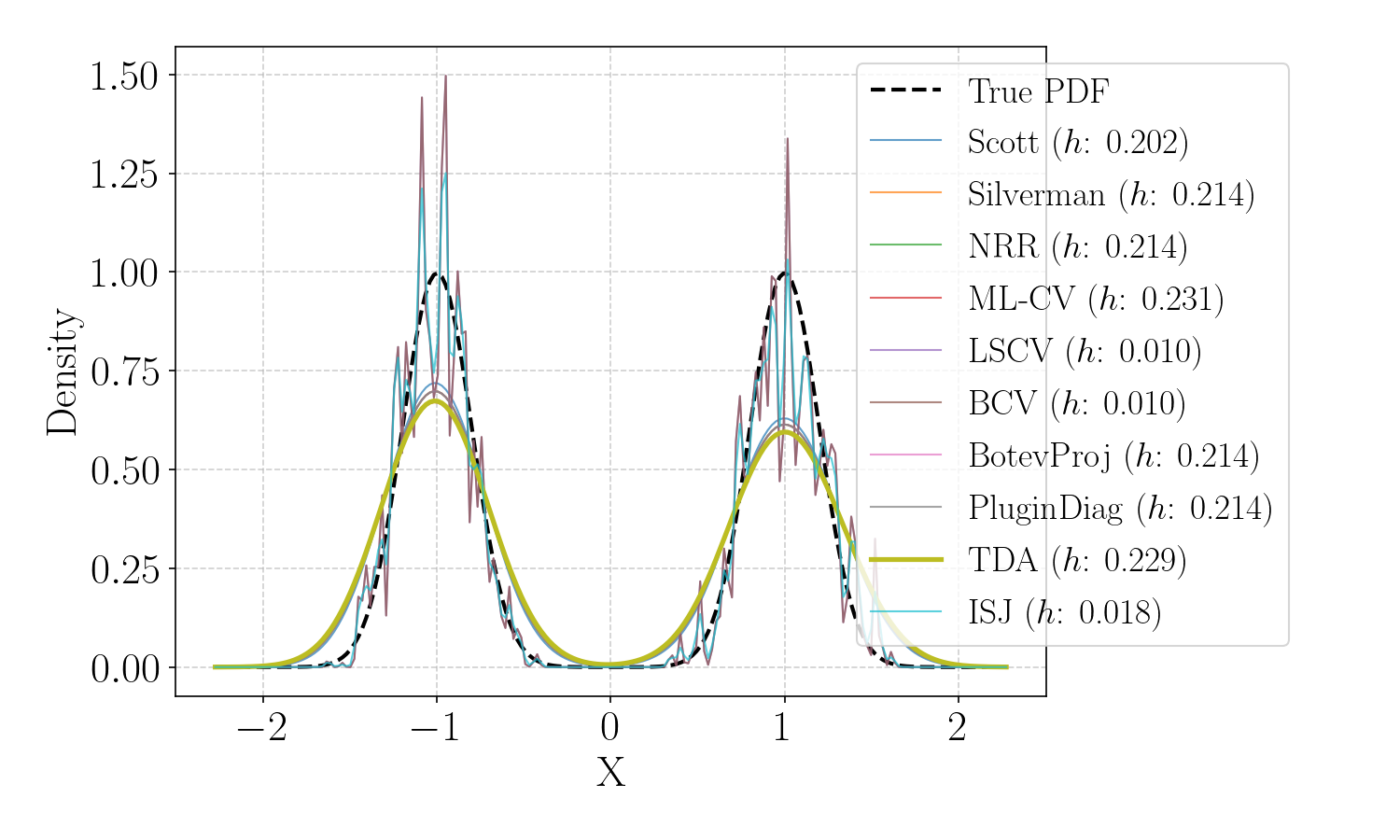}
\caption{The loss landscape, TDA-based KDE and a comparison of all the KDE methods for 1D bimodal distribution.}
\label{fig:1D_KDE1}
\end{figure}

\paragraph{Bimodal Gaussian: }A symmetric mixture of two narrow Gaussian modes centered at -1 and 1, producing a clean bimodal density. See Fig.~\ref{fig:1D_KDE1} for the loss landscape and comparison of all KDEs.
\begin{equation}
p_{\text{bimodal}}(x)
= \mathcal{N}(x;\,-1,\,0.2^2) + \mathcal{N}(x;\,1,\,0.2^2).
\label{eq:1d_1}
\end{equation}

\paragraph{Complex Mixture: }A heterogeneous three-component mixture combining a narrow left Gaussian, a broad central Gaussian, and a sharp Cauchy spike on the right---producing strong multi-scale behavior. 
\begin{equation}
p_{\text{complex}}(x)
= \mathcal{N}(x; -4,\,0.4^2) 
+ \mathcal{N}(x;0,\,1) 
+ 0.2 \, \mathrm{Cauchy}(x;\,6,\,0.1).
\label{eq:1d_2}
\end{equation}

\begin{table}[ht!]
\centering
\scriptsize
\caption{1D datasets: mean $\pm$ std over 500 trials. Best valid values in bold.}
\label{tab:1D}
\begin{tabular}{lcc|cc}
\toprule
\multirow{2}{*}{Method} &
\multicolumn{2}{c|}{1D Bimodal} &
\multicolumn{2}{c}{1D Complex} \\
& KLD $\downarrow$ & EMD $\downarrow$ & KLD $\downarrow$ & EMD $\downarrow$ \\
\midrule
Scott       & 0.0019 $\pm$ 0.0002 & 0.0821 $\pm$ 0.0238 & 0.0187 $\pm$ 0.0088 & 0.2990 $\pm$ 0.1106 \\
Silverman   & 0.0022 $\pm$ 0.0002 & 0.0869 $\pm$ 0.0228 & 0.0205 $\pm$ 0.0095 & 0.3104 $\pm$ 0.1095 \\
NRR         & 0.0022 $\pm$ 0.0002 & 0.0869 $\pm$ 0.0228 & 0.0205 $\pm$ 0.0095 & 0.3104 $\pm$ 0.1095 \\
ML-CV       & 0.0031 $\pm$ 0.0004 & 0.0895 $\pm$ 0.0333 & 0.0099 $\pm$ 0.0244 & 0.2370 $\pm$ 0.1581 \\
LSCV        & 0.0025 $\pm$ 0.0005 & \textbf{0.0520 $\pm$ 0.0312} & 0.0308 $\pm$ 0.0121 & 0.2800 $\pm$ 0.1634 \\
BCV         & 0.0025 $\pm$ 0.0005 & \textbf{0.0520 $\pm$ 0.0312} & 0.0308 $\pm$ 0.0121 & 0.2800 $\pm$ 0.1634 \\
BotevProj   & 0.0022 $\pm$ 0.0002 & 0.0869 $\pm$ 0.0228 & 0.0200 $\pm$ 0.0072 & 0.3089 $\pm$ 0.1066 \\
PluginDiag  & 0.0022 $\pm$ 0.0002 & 0.0870 $\pm$ 0.0228 & 0.0205 $\pm$ 0.0095 & 0.3106 $\pm$ 0.1095 \\
TDA         & \textbf{0.0018 $\pm$ 0.0012} & 0.0776 $\pm$ 0.0319 &
             {0.0149 $\pm$ 0.0065} & {0.2714 $\pm$ 0.1040} \\
ISJ         & 0.0023 $\pm$ 0.0013 & 0.0525 $\pm$ 0.0317 & \textbf{0.0091 $\pm$ 0.0048} & \textbf{0.2101 $\pm$ 0.1274} \\
\bottomrule
\end{tabular}
\end{table}

\subsection{Two-Dimensional Test Densities}

In 2D, we tested a variety of distributions (Eqs.~\ref{eq:2d_1}-\ref{eq:2d_3}). For all, 500 simulations were run with 2000 points sampled in each, with KDE computed on a grid size of $100 \times 100$. 
For the 2D clusters, TDA yields moderate performance: its KLD is somewhat higher than Scott/BotevProj but still far below the highly unstable LSCV/BCV variants. Likewise, its EMD is slightly worse than Scott/BotevProj and similar to ML-CV/LSCV. This indicates that TDA successfully identifies cluster topology but produces bandwidths that are somewhat larger than cluster-optimal. For the elliptical mixture, TDA does reasonably well on this elongated, correlated distribution. Its KLD is close to Scott and ML-CV, outperforming PluginDiag and massively beating LSCV/BCV. The method reliably preserves the orientation and eccentricity of the ellipse, even though it does not outperform classical smoothers that already closely match the Gaussian model. For the weibull dataset with strong skewness. TDA performs quite well. Its KLD is competitive with ML-CV and better than all reference-rule methods and PluginDiag/BotevProj. EMD also aligns with the best-performing methods. Importantly, TDA remains stable and does not exhibit the large variance seen in BCV/LSCV. This suggests topological bandwidth selection handles skewed shapes quite effectively. See Table~\ref{tab:2D} for a quantitative comparison.

\begin{figure}[!htbp]
\centering
\includegraphics[width=1\linewidth]{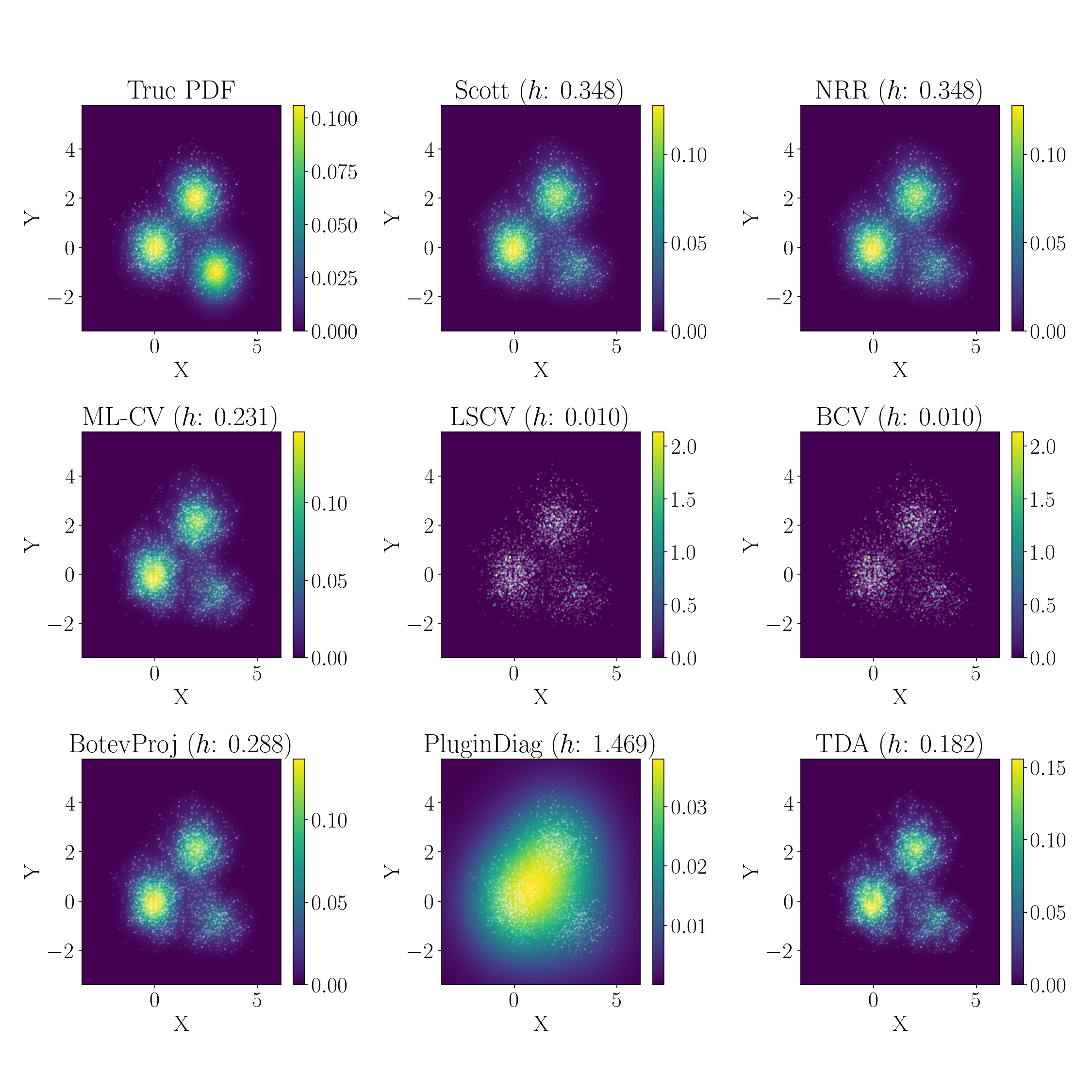}
\caption{A comparative plot of all KDE methods for 2D clusters distribution.}
\label{fig:2D_KDE1}
\end{figure}

\paragraph{Cluster Mixture: }A mixture of three moderately separated Gaussian-like clusters with varying spread and location. See Fig.~\ref{fig:2D_KDE1} for a comparison of all KDEs.
\begin{equation}
p_{\text{clusters}}(x,y)
=\sum_{k=1}^3 
\frac{\exp\!\big(-\tfrac12 (z-\mu_k)^\top \Sigma^{-1} (z-\mu_k)\big)}
{\sqrt{\det\Sigma}},\quad z=(x,y).
\label{eq:2d_1}
\end{equation}

\paragraph{Elliptical Mixture: }A mixture of two strongly anisotropic elliptical clusters, each stretched along different principal axes.
\begin{equation}
p_{\text{elliptical}}(x,y)
= \exp\!\left(-\frac{(x+1)^2}{0.2}-\frac{(y+1)^2}{5}\right)
+ \exp\!\left(-\frac{(x-1)^2}{5}-\frac{(y-1)^2}{0.1^2}\right).
\label{eq:2d_2}
\end{equation}

\paragraph{Weibull: }A radially symmetric distribution with density decreasing according to the Weibull decay.
\begin{equation}
p_{\text{weibull}}(x,y)
= \frac{\alpha}{\beta}\left(\frac{r}{\beta}\right)^{\alpha-1}
\exp\!\left[-\left(\frac{r}{\beta}\right)^{\alpha}\right],
\qquad r=\sqrt{x^{2}+y^{2}}.
\label{eq:2d_3}
\end{equation}

\begin{table}[ht!]
\centering
\scriptsize
\caption{KLD and EMD on 2D datasets. Mean $\pm$ std over 500 trials.}
\label{tab:2D}
\begin{tabular}{lcc|cc|cc}
\toprule
\multirow{2}{*}{Method} &
\multicolumn{2}{c|}{2D Weibull} &
\multicolumn{2}{c|}{2D Elliptical} &
\multicolumn{2}{c}{2D Clusters} \\
& KLD & EMD & KLD & EMD & KLD & EMD \\
\midrule
Scott        & 0.0217$\pm$0.0131 & 32.94$\pm$13.67
             & 0.0063$\pm$0.0622 & 67.82$\pm$38.64
             & 0.0010$\pm$0.0005 & 46.09$\pm$20.75 \\
Silverman    & 0.0217$\pm$0.0131 & 32.94$\pm$13.67
             & 0.0063$\pm$0.0622 & 67.82$\pm$38.64
             & 0.0010$\pm$0.0005 & 46.09$\pm$20.75 \\
NRR          & 0.0217$\pm$0.0131 & 32.94$\pm$13.67
             & 0.0063$\pm$0.0622 & 67.82$\pm$38.64
             & 0.0010$\pm$0.0005 & 46.09$\pm$20.75 \\
ML-CV        & 0.0113$\pm$0.0071 & \textbf{32.46$\pm$13.46}
             & \textbf{0.0060$\pm$0.0487} & 69.56$\pm$37.81
             & 0.0011$\pm$0.0005 & 46.86$\pm$20.45 \\
LSCV         & 0.1073$\pm$0.0395 & 33.83$\pm$12.90
             & 0.0941$\pm$0.1299 & 81.21$\pm$36.95
             & 0.0926$\pm$0.0559 & 58.88$\pm$22.51 \\
BCV          & 1.0385$\pm$0.1666 & 45.19$\pm$11.83
             & 0.0941$\pm$0.1299 & 81.25$\pm$37.00
             & 0.1254$\pm$0.0123 & 63.67$\pm$19.07 \\
BotevProj    & 0.0286$\pm$0.0166 & 33.31$\pm$13.60
             & 0.0067$\pm$0.0771 & \textbf{66.61$\pm$39.01}
             & \textbf{0.0009$\pm$0.0005} & \textbf{45.50$\pm$21.11} \\
PluginDiag   & 0.0116$\pm$0.0053 & 35.75$\pm$12.18
             & 0.0102$\pm$0.0334 & 118.17$\pm$33.84
             & 0.0053$\pm$0.0007 & 96.11$\pm$12.22 \\
TDA          & \textbf{0.0109$\pm$0.0060} & 33.60$\pm$12.95
             & {0.0070$\pm$0.0329} & 82.05$\pm$35.12
             & 0.0027$\pm$0.0012 & 62.93$\pm$21.83 \\
\bottomrule
\end{tabular}
\end{table}

\subsection{Three-Dimensional Test Densities}

In 3D, we tested three different distributions (Eqs.~\ref{eq:3d_1}-\ref{eq:3d_3}). For each, 500 simulations were run with 1000 sampled points, with KDE computed on a grid size of 70 in each dimension. TDA yields the best KLD among all methods for all three distributions---substantially outperforming Scott/Silverman/NRR, ML-CV, LSCV, BotevProj, and PluginDiag. The improvement is large---TDA recovers a smoother, more stable bandwidth in high dimensions, where CV-based methods degrade due to high variance. This clearly demonstrates TDA’s strength in moderately high dimensions. Table~\ref{tab:3D} shows the quantitative outcomes for KLD (EMD has not been calculated due to the high computational cost). For the gaussian dataset, classical rules produce relatively large errors, and CV methods have high variance. Only PlugInDiag is competitive but even compared to that, TDA performs far better. Similarly, for the heavy railed distribution, the gap between the performances widens further---likely because heavy tails tend to confuse global MISE-based or cross-validation-based objectives. Similar results can be seen for the manifold dataset.

\paragraph{Gaussian: }A sharp peaked Gaussian with moderate decay in all dimensions.
\begin{equation}
p(x,y,z)=\exp\!\big(-10[(x^2-h)^2 + (y^2-h)^2 + (z^2-h)^2]\big).
\label{eq:3d_1}
\end{equation}

\paragraph{Heavy-Tailed Mixture: }A mixture of a sharp central spike with a broad Cauchy-like tail, producing heavy-tailed radial structure.
\begin{equation}
p(x,y,z)=0.7\,(1+x^2+y^2+z^2)^{-1} + 
0.3\,\exp\!\big(-500(x^2+y^2+z^2)\big).
\label{eq:3d_2}
\end{equation}

\paragraph{Manifold: }A distribution concentrated near a 1D nonlinear curve embedded in $\mathbb{R}^3$, forming a thin manifold.
\begin{equation}
p(x,y,z)=\exp\!\left(-(y-\sin 5x)^2-(z-\cos 3x)^2\right).
\label{eq:3d_3}
\end{equation}

\begin{table}[ht!]
\centering
\scriptsize
\caption{All 3D datasets (mean $\pm$ std). Best valid KLD bolded.}
\label{tab:3D}
\begin{tabular}{lccc}
\toprule
Method & 3D Gauss & 3D Heavy-Tail & 3D Manifold \\
\midrule
Scott       & 0.3755$\pm$0.9352 & 0.4793$\pm$0.4402 & 0.0066$\pm$0.0025 \\
Silverman   & 0.3859$\pm$0.9526 & 0.5014$\pm$0.4575 & 0.0068$\pm$0.0026 \\
NRR         & 0.3859$\pm$0.9526 & 0.5014$\pm$0.4575 & 0.0068$\pm$0.0026 \\
ML-CV       & 0.4525$\pm$1.2471 & 0.5525$\pm$0.7142 & 0.0045$\pm$0.0018 \\
LSCV        & 0.7495$\pm$1.7219 & 1.4238$\pm$1.0275 & 0.0541$\pm$0.0185 \\
BCV         & 0.7842$\pm$1.7510 & 1.6020$\pm$1.0292 & 0.0541$\pm$0.0185 \\
BotevProj   & 0.5094$\pm$1.1815 & 0.7691$\pm$0.6757 & 0.0100$\pm$0.0039 \\
PluginDiag  & 0.3108$\pm$0.8822 & 0.3186$\pm$0.3918 & 0.0040$\pm$0.0019 \\
TDA         & \textbf{0.2175$\pm$0.7439} &
              \textbf{0.1536$\pm$0.2347} &
              \textbf{0.0039$\pm$0.0018} \\
\bottomrule
\end{tabular}
\end{table}

\subsection{Four-Dimensional Test Densities}

In 4D, we test multiple distributions: a smooth unimodal Gaussian, a multi-scale heavy mixture, and an extremely heavy-tailed density with a sharp central spike (Eqs.~\ref{eq:4d_1}-\ref{eq:4d_3}). For each, 500 simulations were run with 1000 points sampled in each, with KDE computed on a small grid size of 40 in each dimension. The results in Table~\ref{tab:4D} show that TDA continues to scale gracefully with dimension, consistently achieving the lowest KLD among all valid methods---much better than Scott/Silverman/NRR/ML-CV/BotevProj and dramatically better than BCV/LSCV. Classical rules have rapidly increasing variance, producing very large KLD. CV methods also have degraded performance, exhibiting unstable smoothing. For the gaussian dataset, PlugInDiag is the only competitive classical method but TDA has higher accuracy, For the heavy mixture, which is challenging due to its multiple scales, all classical methods return very large KLDs. Even PlugInDiag returns a high KLD compared to TDA. This shows TDA's ability to preserve both fine-scale peaks and large-scale geometry. Likewise, the heavy tail with spike dataset shows similar results with classical rules and CV performing poorly, PlugInDiag somewhat better than them but still getting outperformed by TDA. 

\paragraph{Gaussian: }A smooth, moderately concentrated Gaussian density with no heavy tails.
\begin{equation}
p(x,y,z,w)
=\exp\!\left(-\left[(x^2-h)^2 + (y^2-h)^2 + (z^2-h)^2 + (w^2-h)^2\right]\right).
\label{eq:4d_1}
\end{equation}

\paragraph{Heavy Mixture: }A mixture of two localized Gaussian peaks with a heavy-tailed component, producing multi-scale structure.
\begin{align}
p(x,y,z,w)
&= 0.6\exp\!\left(-\frac{(x-2)^2+(y-2)^2+(z-2)^2+(w-2)^2}{0.1}\right) \nonumber \\
&\quad +0.4\exp\!\left(-\frac{(x+2)^2+(y+2)^2+(z+2)^2+(w+2)^2}{0.1}\right) \nonumber \\
&\quad +0.3\,(1+x^2+y^2+z^2+w^2)^{-1}.
\label{eq:4d_2}
\end{align}

\paragraph{Heavytail with Spike: }A 4D analogue of a sharp central spike mixed with a broad Cauchy-like tail, yielding both high concentration and extreme heavy-tailed behavior.
\begin{equation}
p(x,y,z,w)
= 0.3\,\exp\!\Bigl(-800\,(x^{2}+y^{2}+z^{2}+w^{2})\Bigr)
\;+\;
0.7\,\frac{1}{1 + x^{2}+y^{2}+z^{2}+w^{2}}.
\label{eq:4d_3}
\end{equation}

\begin{table}[ht!]
\centering
\scriptsize
\caption{All 4D datasets (mean $\pm$ std). Best valid KLD bolded.}
\label{tab:4D}
\begin{tabular}{lccc}
\toprule
Method & 4D Gauss & 4D Heavy Mixture & 4D Heavy-Tail \\
\midrule
Scott       & 0.4777$\pm$2.5875 & 3.3190$\pm$2.6269 & 3.3231$\pm$2.6327 \\
Silverman   & 0.4959$\pm$2.6555 & 3.5210$\pm$2.7475 & 3.5251$\pm$2.7535 \\
NRR         & 0.4959$\pm$2.6555 & 3.5210$\pm$2.7475 & 3.5251$\pm$2.7535 \\
ML-CV       & 0.5615$\pm$3.0036 & 4.1814$\pm$3.8767 & 4.1860$\pm$3.8844 \\
LSCV        & 0.6841$\pm$3.2516 & 6.5351$\pm$4.2593 & 6.5358$\pm$4.2655 \\
BCV         & 0.6953$\pm$3.2631 & 6.6055$\pm$4.2653 & 6.6064$\pm$4.2715 \\
BotevProj   & 0.6159$\pm$3.0957 & 5.1073$\pm$3.7108 & 5.1088$\pm$3.7162 \\
PluginDiag  & 0.4600$\pm$2.6133 & 2.9982$\pm$2.8766 & 3.0027$\pm$2.8831 \\
TDA         & \textbf{0.2754$\pm$2.0012} &
              \textbf{1.1352$\pm$1.8704} &
              \textbf{1.1383$\pm$1.8745} \\
\bottomrule
\end{tabular}
\end{table}

\subsection{Real-Data: MNIST}

\begin{figure}[!htbp]
\centering
\includegraphics[width=1\linewidth]{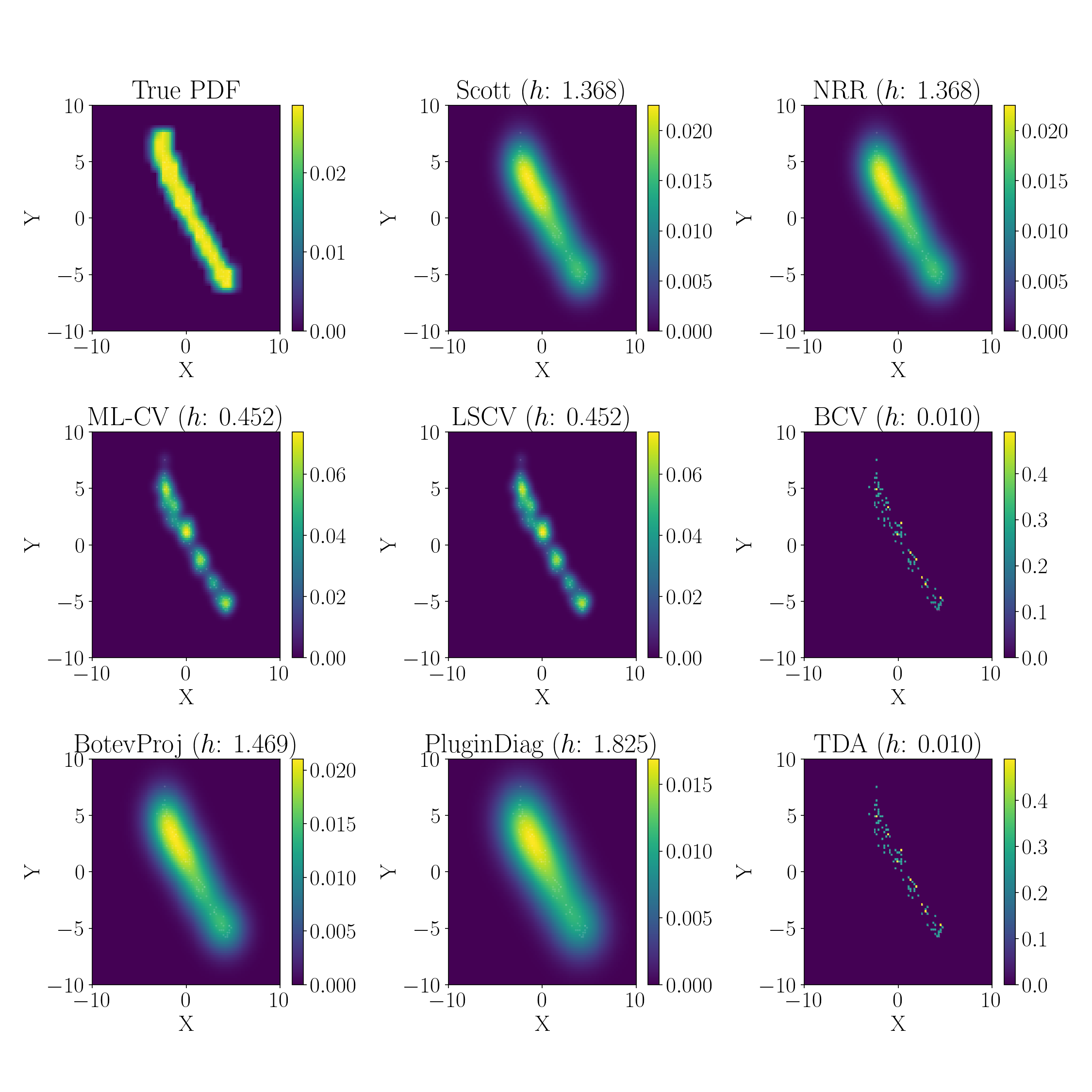}
\caption{A comparative plot of all KDE methods for digit 1.}
\label{fig:mnist}
\end{figure}

To evaluate the performance of our method on
real-world data, we conducted experiments on the MNIST handwritten-digit dataset. Since results are qualitatively similar across all digits $0$-$9$, we follow
common practice and report values for representative digits:
\textit{1} (simple), \textit{3} (half loops), and \textit{8} (full loops). These digits span the spectrum of MNIST density variability and are often used as canonical representatives. For each digit class, we treat the empirical distribution of all training images as a reference density on a normalized pixel grid, and sample a point cloud from them. Table~\ref{tab:mnist} summarizes the performance of all bandwidth selectors. On many digits, TDA achieves EMD close to the best classical methods, and its variability is moderate. KLD tends to be slightly worse than Scott/Silverman/BotevProj, reflecting oversmoothing from fixed-epoch training. However, TDA never catastrophically fails, unlike BCV---and has overlapping error bars with the best performing methods in each case. TDA is competitive on most digits in terms of shape preservation (EMD), though its KLD sometimes increases due to edge blurring. See Figure~\ref{fig:mnist} for a comparative plot of all KDEs for the digit 1. 

Additionally, we also evaluated the computational cost of each bandwidth selection method during this experiment. As expected, classical rules such as Scott, Silverman, and NRR execute almost instantaneously (on the order of $10^{-4}$ seconds), while cross-validation methods (ML-CV, LSCV, BCV) incur moderate overhead due to repeated density evaluations. Botev’s projection estimator and the diagonal plug-in method also remain efficient, typically running within a few milliseconds. In contrast, the TDA-based bandwidth selector is slower, with average runtimes around 18 seconds for a single digit image. This disparity is not inherent to the loss formulation itself but to the use of a fixed, conservative number of optimization epochs---intentionally kept far larger than necessary for sure-convergence. Incorporating adaptive stopping criteria or dynamic learning-rate schedules is expected to dramatically reduce runtime.

\begin{table}[ht!]
\centering
\scriptsize
\caption{MNIST digits 1, 3, and 8. Mean $\pm$ std over 50 trials. Best valid values are bolded.}
\label{tab:mnist}
\begin{tabular}{lcc|cc|cc}
\toprule
\multirow{2}{*}{Method} &
\multicolumn{2}{c|}{Digit 1} &
\multicolumn{2}{c|}{Digit 3} &
\multicolumn{2}{c}{Digit 8} \\
& KLD $\downarrow$ & EMD $\downarrow$ & KLD $\downarrow$ & EMD $\downarrow$ & KLD $\downarrow$ & EMD $\downarrow$ \\
\midrule
Scott        & 0.0175$\pm$0.0012 & 20.45$\pm$3.84 
             & 0.0174$\pm$0.0010 & 24.37$\pm$6.43
             & 0.0130$\pm$0.0006 & 18.18$\pm$3.33 \\
Silverman    & 0.0175$\pm$0.0012 & 20.45$\pm$3.84 
             & 0.0174$\pm$0.0010 & 24.37$\pm$6.43
             & 0.0130$\pm$0.0006 & 18.18$\pm$3.33 \\
NRR          & 0.0175$\pm$0.0012 & 20.45$\pm$3.84 
             & 0.0174$\pm$0.0010 & 24.37$\pm$6.43
             & 0.0130$\pm$0.0006 & 18.18$\pm$3.33 \\
ML-CV        & \textbf{0.0117$\pm$0.0126} & \textbf{13.76$\pm$5.15}
             & 0.0165$\pm$0.0058 & \textbf{22.07$\pm$7.13}
             & 0.0144$\pm$0.0048 & 18.55$\pm$4.01 \\
LSCV         & 0.0297$\pm$0.0267 & 14.44$\pm$5.14
             & 0.0177$\pm$0.0051 & 22.22$\pm$7.16
             & 0.0186$\pm$0.0228 & 18.70$\pm$4.16 \\
BCV          & 0.8603$\pm$0.0037 & 17.65$\pm$4.19
             & 0.8846$\pm$0.0009 & 26.58$\pm$6.36
             & 0.8793$\pm$0.0015 & 22.88$\pm$3.51 \\
BotevProj    & 0.0156$\pm$0.0040 & 19.16$\pm$5.14
             & \textbf{0.0156$\pm$0.0022} & 23.49$\pm$6.71
             & \textbf{0.0109$\pm$0.0021} & \textbf{17.69$\pm$3.63} \\
PluginDiag   & 0.0261$\pm$0.0008 & 27.44$\pm$2.98
             & 0.0247$\pm$0.0006 & 29.72$\pm$5.30
             & 0.0201$\pm$0.0007 & 23.01$\pm$2.48 \\
TDA          & 0.0232$\pm$0.0499 & 15.85$\pm$9.60
             & 0.0172$\pm$0.0044 & 24.34$\pm$7.40
             & 0.0157$\pm$0.0020 & 19.47$\pm$3.38 \\
\bottomrule
\end{tabular}
\end{table}

\section{Limitations}

The primary limitation of the proposed approach lies in the cost of computing cubical persistent homology in dimensions above four. For a grid of size $n$ in $d$ dimensions, the worst-case complexity of the standard cubical persistence algorithm~\citet{CubicalComplex} is
\[
O(d3^{d} n + d^{2} 2^{d} n),
\]
as established in~\citet{Wagner2011}.  
Since the KDE must be evaluated on a full $d$-dimensional grid and the resulting cubical complex must be processed by the persistence algorithm, the computational burden grows rapidly with dimension. In practice, this may be overcome by downsampling or slicing strategies, or through use of recent work such as the divide-and-conquer persistent homology algorithm~\cite{DAC} which offers speedups through domain decomposition and parallelization. Alternatively, instead of grids, using Vietoris-Rips persistence form of loss, directly on the points, can be considered---which would allow for higher-dimensional analysis easily.

\section{Future Work}

There are several natural extensions of this work. First, while this work focuses on a single global bandwidth, the topological loss formulation can generalize to \emph{anisotropic} or \emph{dimension-wise} bandwidths. Allowing the optimizer to choose a separate bandwidth per coordinate, or even a full bandwidth matrix, could significantly improve performance on elongated or correlated densities.
Second, although the current method is kernel-agnostic, all experiments fixed a gaussian kernel during optimization. A promising direction is to jointly learn both the bandwidth and the kernel shape, either from a parametric dictionary (Gaussian, Epanechnikov, Laplace, etc.) or via a learned mixture-of-kernels representation. This would move toward a fully data-adaptive, topology-regularized formulation of density estimation in which both the scale and geometry of smoothing are optimized to preserve the correct homological structure of the underlying distribution. Finally, a natural avenue for future work is to develop a deeper statistical theory for the proposed bandwidth selection method. Although empirical results demonstrate strong performance across a wide range of densities, several theoretical questions remain open for future investigations.  

\section{Conclusion}

This work introduces a topology-driven framework for automated bandwidth selection in kernel density estimation, offering an unsupervised alternative to classical rules and cross-validation methods. Rather than relying on parametric assumptions, our method selects the bandwidth by optimizing a topological loss function which encourages density estimates that preserve salient structural features of the distribution. The TDA-guided bandwidth often matches or exceeds the performance of standard selectors across a broad suite of synthetic datasets ranging from one to four dimensions, particularly in settings where the underlying density exhibits non-Gaussian, multi-modal, or heavy-tailed behavior. Importantly, the method is very effective in high dimensions where visualization is impossible and structural inference must rely purely on quantitative criteria. To assess real-world applicability, we additionally evaluated the method on MNIST digit distributions. The topology-guided bandwidth produced competitive estimates, highlighting the promise of topological information as a practical guide for smoothing parameter selection. 
The code can be accessed via \url{https://github.com/stanweer1/Unsupervised-Learning-of-Density-Estimates-with-Topological-Optimization}.

\begin{ack}
This material is based upon work supported by the Air Force Office of Scientific Research under award number FA9550-26-1-0011. S.T.~thanks Dr.~Maxwell Chumley for insight into topological optimization, and acknowledges the funding from NSF Frontera Computational Science Fellowship award 2025-2026. 
\end{ack}

{
\small
\bibliography{bibfile}

\begin{thebibliography}{37}
\providecommand{\natexlab}[1]{#1}
\providecommand{\url}[1]{\texttt{#1}}
\expandafter\ifx\csname urlstyle\endcsname\relax
  \providecommand{\doi}[1]{doi: #1}\else
  \providecommand{\doi}{doi: \begingroup \urlstyle{rm}\Url}\fi

\bibitem[Abramson(1982)]{Abramson1982}
Ian~S. Abramson.
\newblock On bandwidth variation in kernel estimates-a square root law.
\newblock \emph{The Annals of Statistics}, 10\penalty0 (4), December 1982.
\newblock ISSN 0090-5364.
\newblock \doi{10.1214/aos/1176345986}.
\newblock URL \url{http://dx.doi.org/10.1214/AOS/1176345986}.

\bibitem[Chazal et~al.(2018)Chazal, Fasy, Lecci, Bertr, Michel, Aless, ro~Rinaldo, and Wasserman]{Fasy2018}
Fr{\'e}d{\'e}ric Chazal, Brittany Fasy, Fabrizio Lecci, Bertr, Michel, Aless, ro~Rinaldo, and Larry Wasserman.
\newblock Robust topological inference: Distance to a measure and kernel distance.
\newblock \emph{Journal of Machine Learning Research}, 18\penalty0 (159):\penalty0 1--40, 2018.
\newblock URL \url{http://jmlr.org/papers/v18/15-484.html}.

\bibitem[Chen(2018)]{Chen2018}
Yen‐Chi Chen.
\newblock Modal regression using kernel density estimation: A review.
\newblock \emph{WIREs Computational Statistics}, 10\penalty0 (4), February 2018.
\newblock ISSN 1939-0068.
\newblock \doi{10.1002/wics.1431}.
\newblock URL \url{http://dx.doi.org/10.1002/wics.1431}.

\bibitem[Cheng et~al.(2018)Cheng, Gao, and Zhang]{Cheng2018}
Tingting Cheng, Jiti Gao, and Xibin Zhang.
\newblock Nonparametric localized bandwidth selection for kernel density estimation.
\newblock \emph{Econometric Reviews}, 38\penalty0 (7):\penalty0 733–762, February 2018.
\newblock ISSN 1532-4168.
\newblock \doi{10.1080/07474938.2017.1397835}.
\newblock URL \url{http://dx.doi.org/10.1080/07474938.2017.1397835}.

\bibitem[Cho et~al.(2020)Cho, Hwang, and Suh]{Cho2020}
Jaewoong Cho, Gyeongjo Hwang, and Changho Suh.
\newblock A fair classifier using kernel density estimation.
\newblock In H.~Larochelle, M.~Ranzato, R.~Hadsell, M.F. Balcan, and H.~Lin, editors, \emph{Advances in Neural Information Processing Systems}, volume~33, pages 15088--15099. Curran Associates, Inc., 2020.
\newblock URL \url{https://proceedings.neurips.cc/paper_files/paper/2020/file/ac3870fcad1cfc367825cda0101eee62-Paper.pdf}.

\bibitem[Cohen-Steiner et~al.(2006)Cohen-Steiner, Edelsbrunner, and Harer]{CohenSteiner2006}
David Cohen-Steiner, Herbert Edelsbrunner, and John Harer.
\newblock Stability of persistence diagrams.
\newblock \emph{Discrete \& Computational Geometry}, 37\penalty0 (1):\penalty0 103–120, December 2006.
\newblock ISSN 1432-0444.
\newblock \doi{10.1007/s00454-006-1276-5}.
\newblock URL \url{http://dx.doi.org/10.1007/s00454-006-1276-5}.

\bibitem[Davies and Baddeley(2017)]{Davies2017}
Tilman~M. Davies and Adrian Baddeley.
\newblock Fast computation of spatially adaptive kernel estimates.
\newblock \emph{Statistics and Computing}, 28\penalty0 (4):\penalty0 937–956, August 2017.
\newblock ISSN 1573-1375.
\newblock \doi{10.1007/s11222-017-9772-4}.
\newblock URL \url{http://dx.doi.org/10.1007/s11222-017-9772-4}.

\bibitem[Delaigle and Gijbels(2004)]{Delaigle2004}
A.~Delaigle and I.~Gijbels.
\newblock Practical bandwidth selection in deconvolution kernel density estimation.
\newblock \emph{Computational Statistics and Data Analysis}, 45\penalty0 (2):\penalty0 249–267, March 2004.
\newblock ISSN 0167-9473.
\newblock \doi{10.1016/s0167-9473(02)00329-8}.
\newblock URL \url{http://dx.doi.org/10.1016/S0167-9473(02)00329-8}.

\bibitem[Dey and Wang(2022)]{TamalDey2022}
Tamal~Krishna Dey and Yusu Wang.
\newblock \emph{Computational topology for data analysis}.
\newblock Cambridge University Press, Cambridge, England, mar 2022.

\bibitem[Dlotko(2025)]{CubicalComplex}
Pawel Dlotko.
\newblock Cubical complex.
\newblock In \emph{GUDHI User and Reference Manual}. GUDHI Editorial Board, 3.11.0 edition, 2025.
\newblock URL \url{https://gudhi.inria.fr/doc/3.11.0/group__cubical__complex.html}.

\bibitem[Dtchetgnia~Djeundam et~al.(2013)Dtchetgnia~Djeundam, Yamapi, Kofane, and Aziz-Alaoui]{DtchetgniaDjeundam2013}
S.~R. Dtchetgnia~Djeundam, R.~Yamapi, T.~C. Kofane, and M.~A. Aziz-Alaoui.
\newblock Deterministic and stochastic bifurcations in the hindmarsh-rose neuronal model.
\newblock \emph{Chaos: An Interdisciplinary Journal of Nonlinear Science}, 23\penalty0 (3), August 2013.
\newblock ISSN 1089-7682.
\newblock \doi{10.1063/1.4818545}.
\newblock URL \url{http://dx.doi.org/10.1063/1.4818545}.

\bibitem[Duin(1976)]{Duin1976}
Duin.
\newblock On the choice of smoothing parameters for parzen estimators of probability density functions.
\newblock \emph{IEEE Transactions on Computers}, C–25\penalty0 (11):\penalty0 1175–1179, November 1976.
\newblock ISSN 2326-3814.
\newblock \doi{10.1109/tc.1976.1674577}.
\newblock URL \url{http://dx.doi.org/10.1109/TC.1976.1674577}.

\bibitem[Duong and Hazelton(2005)]{DUONG2005}
Tarn Duong and Martin~L. Hazelton.
\newblock Cross-validation bandwidth matrices for multivariate kernel density estimation.
\newblock \emph{Scandinavian Journal of Statistics}, 32\penalty0 (3):\penalty0 485–506, September 2005.
\newblock ISSN 1467-9469.
\newblock \doi{10.1111/j.1467-9469.2005.00445.x}.
\newblock URL \url{http://dx.doi.org/10.1111/j.1467-9469.2005.00445.x}.

\bibitem[Gao et~al.(2022)Gao, Jiang, and Qian]{Gao2022}
Jia-Xing Gao, Da-Quan Jiang, and Min-Ping Qian.
\newblock Adaptive manifold density estimation.
\newblock \emph{Journal of Statistical Computation and Simulation}, 92\penalty0 (11):\penalty0 2317–2331, January 2022.
\newblock ISSN 1563-5163.
\newblock \doi{10.1080/00949655.2022.2028283}.
\newblock URL \url{http://dx.doi.org/10.1080/00949655.2022.2028283}.

\bibitem[Ghosh et~al.(2006)Ghosh, Chaudhuri, and Sengupta]{Ghosh2006}
Anil~K Ghosh, Probal Chaudhuri, and Debasis Sengupta.
\newblock Classification using kernel density estimates: Multiscale analysis and visualization.
\newblock \emph{Technometrics}, 48\penalty0 (1):\penalty0 120–132, February 2006.
\newblock ISSN 1537-2723.
\newblock \doi{10.1198/004017005000000391}.
\newblock URL \url{http://dx.doi.org/10.1198/004017005000000391}.

\bibitem[Gramacki and Gramacki(2015)]{Botev}
Artur Gramacki and Jarosław Gramacki.
\newblock Fft-based fast bandwidth selector for multivariate kernel density estimation.
\newblock \emph{arXiv}, 2015.
\newblock \doi{10.48550/ARXIV.1511.07482}.
\newblock URL \url{https://arxiv.org/abs/1511.07482}.

\bibitem[Hall et~al.(1991)Hall, Sheather, Jones, and Marron]{HALL1991}
Peter Hall, Simon~J. Sheather, M.~C. Jones, and J.~S. Marron.
\newblock On optimal data-based bandwidth selection in kernel density estimation.
\newblock \emph{Biometrika}, 78\penalty0 (2):\penalty0 263–269, 1991.
\newblock ISSN 1464-3510.
\newblock \doi{10.1093/biomet/78.2.263}.
\newblock URL \url{http://dx.doi.org/10.1093/biomet/78.2.263}.

\bibitem[Han et~al.(2021)Han, Wang, Xu, Zhang, and Ren]{Han2021}
Ping Han, Liang Wang, Wei Xu, Hongxia Zhang, and Zhicong Ren.
\newblock The stochastic p-bifurcation analysis of the impact system via the most probable response.
\newblock \emph{Chaos, Solitons and Fractals}, 144:\penalty0 110631, March 2021.
\newblock ISSN 0960-0779.
\newblock \doi{10.1016/j.chaos.2020.110631}.
\newblock URL \url{http://dx.doi.org/10.1016/j.chaos.2020.110631}.

\bibitem[Hu et~al.(2020)Hu, Gao, Li, Wu, Du, and Maybank]{Hu2020}
Weiming Hu, Jun Gao, Bing Li, Ou~Wu, Junping Du, and Stephen Maybank.
\newblock Anomaly detection using local kernel density estimation and context-based regression.
\newblock \emph{IEEE Transactions on Knowledge and Data Engineering}, 32\penalty0 (2):\penalty0 218–233, February 2020.
\newblock ISSN 2326-3865.
\newblock \doi{10.1109/tkde.2018.2882404}.
\newblock URL \url{http://dx.doi.org/10.1109/TKDE.2018.2882404}.

\bibitem[Huntsman(2017)]{TDE}
Steve Huntsman.
\newblock Topological density estimation.
\newblock \emph{arXiv}, 2017.
\newblock \doi{10.48550/ARXIV.1701.09025}.
\newblock URL \url{https://arxiv.org/abs/1701.09025}.

\bibitem[Jones et~al.(1996)Jones, Marron, and Sheather]{Jones1996}
M.~C. Jones, J.~S. Marron, and S.~J. Sheather.
\newblock A brief survey of bandwidth selection for density estimation.
\newblock \emph{Journal of the American Statistical Association}, 91\penalty0 (433):\penalty0 401–407, March 1996.
\newblock ISSN 1537-274X.
\newblock \doi{10.1080/01621459.1996.10476701}.
\newblock URL \url{http://dx.doi.org/10.1080/01621459.1996.10476701}.

\bibitem[Kan et~al.(2025)Kan, Wang, Zhang, and Niu]{Kan2025}
Zijian Kan, Jun Wang, Jianchao Zhang, and Jiangchuan Niu.
\newblock Random p-bifurcation in a duffing–van der pol vibro-impact system with a bingham model.
\newblock \emph{Chaos: An Interdisciplinary Journal of Nonlinear Science}, 35\penalty0 (2), February 2025.
\newblock ISSN 1089-7682.
\newblock \doi{10.1063/5.0246296}.
\newblock URL \url{http://dx.doi.org/10.1063/5.0246296}.

\bibitem[Kumar et~al.(2016)Kumar, Narayanan, and Gupta]{Kumar2016}
Pankaj Kumar, S.~Narayanan, and Sayan Gupta.
\newblock Stochastic bifurcations in a vibro-impact duffing–van der pol oscillator.
\newblock \emph{Nonlinear Dynamics}, 85\penalty0 (1):\penalty0 439–452, March 2016.
\newblock ISSN 1573-269X.
\newblock \doi{10.1007/s11071-016-2697-1}.
\newblock URL \url{http://dx.doi.org/10.1007/s11071-016-2697-1}.

\bibitem[Laing and Lord(2009)]{StochMethods2009}
Carlo Laing and Gabriel~J Lord.
\newblock \emph{Stochastic Methods in Neuroscience}.
\newblock Oxford University Press, September 2009.
\newblock ISBN 9780199235070.
\newblock \doi{10.1093/acprof:oso/9780199235070.001.0001}.
\newblock URL \url{http://dx.doi.org/10.1093/acprof:oso/9780199235070.001.0001}.

\bibitem[Li and Cisewski-Kehe(2024)]{DAC}
Chenghui Li and Jessi Cisewski-Kehe.
\newblock A divide-and-conquer approach to persistent homology.
\newblock \emph{arXiv}, 2024.
\newblock \doi{10.48550/ARXIV.2410.01839}.
\newblock URL \url{https://arxiv.org/abs/2410.01839}.

\bibitem[Mousavinejad et~al.(2022)Mousavinejad, FatehiNia, and Ebrahimi]{Mousavinejad2022}
F.~S. Mousavinejad, M.~FatehiNia, and A.~Ebrahimi.
\newblock P-bifurcation of stochastic van der pol model as a dynamical system in neuroscience.
\newblock \emph{Communications on Applied Mathematics and Computation}, 4\penalty0 (4):\penalty0 1293–1312, March 2022.
\newblock ISSN 2661-8893.
\newblock \doi{10.1007/s42967-021-00176-9}.
\newblock URL \url{http://dx.doi.org/10.1007/s42967-021-00176-9}.

\bibitem[Mugdadi and Ahmad(2004)]{Mugdadi2004}
A.R Mugdadi and Ibrahim~A Ahmad.
\newblock A bandwidth selection for kernel density estimation of functions of random variables.
\newblock \emph{Computational Statistics and Data Analysis}, 47\penalty0 (1):\penalty0 49–62, August 2004.
\newblock ISSN 0167-9473.
\newblock \doi{10.1016/j.csda.2003.10.013}.
\newblock URL \url{http://dx.doi.org/10.1016/j.csda.2003.10.013}.

\bibitem[Qiao(2020)]{Qiao2020}
Wanli Qiao.
\newblock Asymptotics and optimal bandwidth for nonparametric estimation of density level sets.
\newblock \emph{Electronic Journal of Statistics}, 14\penalty0 (1), January 2020.
\newblock ISSN 1935-7524.
\newblock \doi{10.1214/19-ejs1668}.
\newblock URL \url{http://dx.doi.org/10.1214/19-EJS1668}.

\bibitem[Scikit-Learn(2020)]{scikitlearnKernelDensity}
Scikit-Learn.
\newblock {K}ernel{D}ensity --- scikit-learn.org.
\newblock \url{https://scikit-learn.org/stable/modules/generated/sklearn.neighbors.KernelDensity.html#sklearn.neighbors.KernelDensity}, 2020.
\newblock [Accessed 11-05-2025].

\bibitem[Scott(2015)]{Scott2015}
David~W. Scott.
\newblock \emph{Multivariate Density Estimation: Theory, Practice, and Visualization}.
\newblock Wiley, March 2015.
\newblock ISBN 9781118575574.
\newblock \doi{10.1002/9781118575574}.
\newblock URL \url{http://dx.doi.org/10.1002/9781118575574}.

\bibitem[Sheather(2004)]{Sheather2004}
Simon~J. Sheather.
\newblock Density estimation.
\newblock \emph{Statistical Science}, 19\penalty0 (4), November 2004.
\newblock ISSN 0883-4237.
\newblock \doi{10.1214/088342304000000297}.
\newblock URL \url{http://dx.doi.org/10.1214/088342304000000297}.

\bibitem[Silverman(2018)]{Silverman2018}
B.W. Silverman.
\newblock \emph{Density Estimation for Statistics and Data Analysis}.
\newblock Routledge, February 2018.
\newblock ISBN 9781315140919.
\newblock \doi{10.1201/9781315140919}.
\newblock URL \url{http://dx.doi.org/10.1201/9781315140919}.

\bibitem[Tanweer et~al.(2024)Tanweer, A.~Khasawneh, Munch, and R.~Tempelman]{Tanweer2024}
Sunia Tanweer, Firas A.~Khasawneh, Elizabeth Munch, and Joshua R.~Tempelman.
\newblock A topological framework for identifying phenomenological bifurcations in stochastic dynamical systems.
\newblock \emph{Nonlinear Dynamics}, 112\penalty0 (6):\penalty0 4687–4703, February 2024.
\newblock ISSN 1573-269X.
\newblock \doi{10.1007/s11071-024-09289-1}.
\newblock URL \url{http://dx.doi.org/10.1007/s11071-024-09289-1}.

\bibitem[Wagner et~al.(2011)Wagner, Chen, and Vu\c{c}ini]{Wagner2011}
Hubert Wagner, Chao Chen, and Erald Vu\c{c}ini.
\newblock \emph{Efficient Computation of Persistent Homology for Cubical Data}, page 91–106.
\newblock Springer Berlin Heidelberg, November 2011.
\newblock ISBN 9783642231759.
\newblock \doi{10.1007/978-3-642-23175-9_7}.
\newblock URL \url{http://dx.doi.org/10.1007/978-3-642-23175-9_7}.

\bibitem[Wang et~al.(2018)Wang, Tsokos, and Saghafi]{Wang2018}
Xing Wang, Chris~P. Tsokos, and Abolfazl Saghafi.
\newblock Improved parameter estimation of time dependent kernel density by using artificial neural networks.
\newblock \emph{The Journal of Finance and Data Science}, 4\penalty0 (3):\penalty0 172–182, September 2018.
\newblock ISSN 2405-9188.
\newblock \doi{10.1016/j.jfds.2018.04.002}.
\newblock URL \url{http://dx.doi.org/10.1016/j.jfds.2018.04.002}.

\bibitem[Zhang et~al.(2006)Zhang, King, and Hyndman]{Zhang2006}
Xibin Zhang, Maxwell~L. King, and Rob~J. Hyndman.
\newblock A bayesian approach to bandwidth selection for multivariate kernel density estimation.
\newblock \emph{Computational Statistics and Data Analysis}, 50\penalty0 (11):\penalty0 3009–3031, July 2006.
\newblock ISSN 0167-9473.
\newblock \doi{10.1016/j.csda.2005.06.019}.
\newblock URL \url{http://dx.doi.org/10.1016/j.csda.2005.06.019}.

\bibitem[Zámečník et~al.(2023)Zámečník, Horová, Katina, and Hasilová]{Zmenk2023}
Stanislav Zámečník, Ivana Horová, Stanislav Katina, and Kamila Hasilová.
\newblock An adaptive method for bandwidth selection in circular kernel density estimation.
\newblock \emph{Computational Statistics}, 39\penalty0 (4):\penalty0 1709–1728, September 2023.
\newblock ISSN 1613-9658.
\newblock \doi{10.1007/s00180-023-01401-0}.
\newblock URL \url{http://dx.doi.org/10.1007/s00180-023-01401-0}.

\end{thebibliography}
}

\appendix
\section{Hyperparameter Sensitivity Study}
\label{sec:sensitivity}

\begin{figure}[!htbp]
\centering
\includegraphics[width=0.45\linewidth]{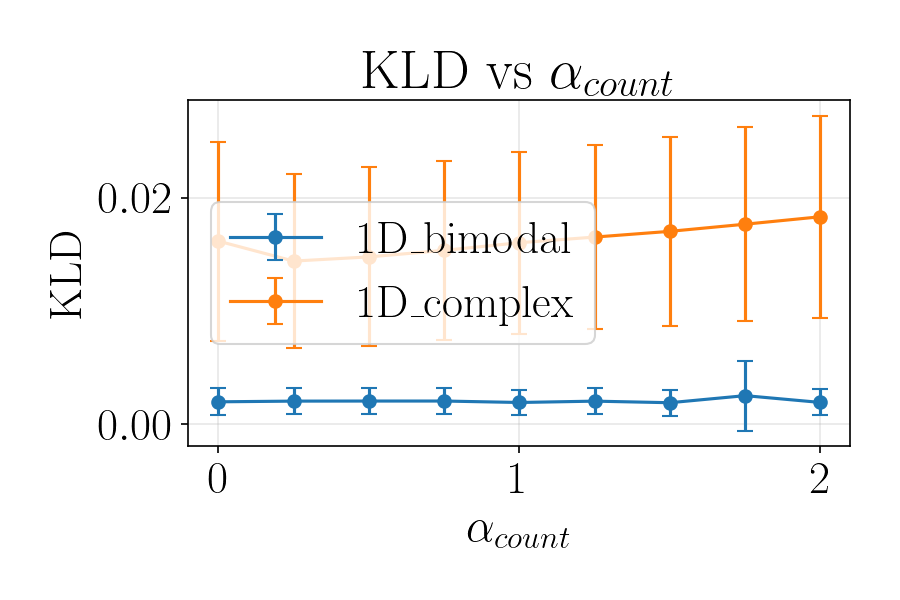}
\includegraphics[width=0.45\linewidth]{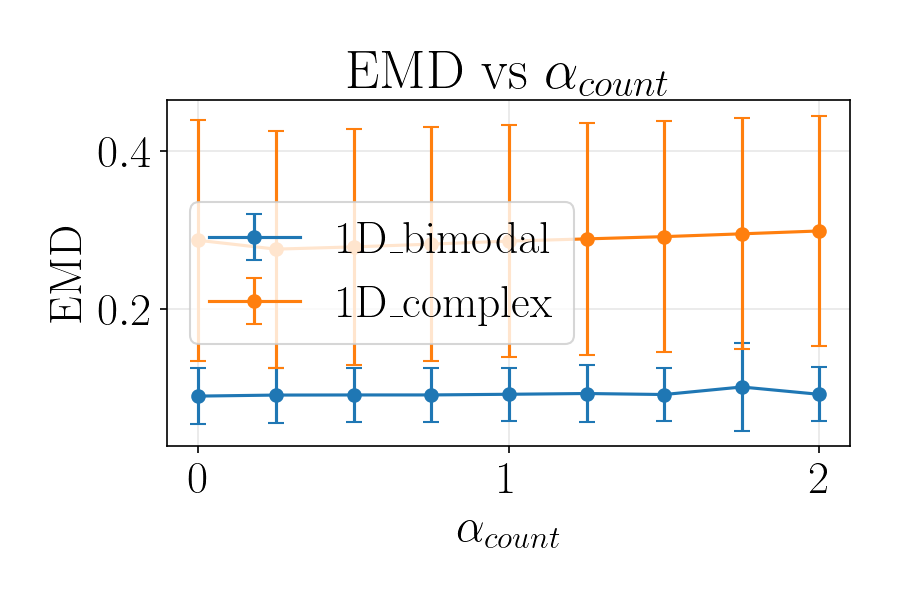}
\caption{KLD and EMD as $\alpha_\mathrm{count}$ is varied in $[0, 2]$.}
\label{fig:alpha_pe}
\end{figure}

To evaluate the robustness of our topology-guided KDE bandwidth selection method, 
we perform a hyperparameter sensitivity analysis on both one-dimensional datasets. For each, the method was run over 25 random seeds, sweeping each hyperparameter independently over a broad range, keeping the remaining parameters fixed at their baseline values. The default configuration used is $\alpha_{\mathrm{count}} = \alpha_{\mathrm{TP}} = 1$ with a \text{grid resolution} of 200. For each configuration we computed both the KLD and EMD between the estimated and true densities.
\paragraph{Sensitivity to \(\alpha_{\mathrm{count}}\):}
In both datasets, the performance varies smoothly and only mildly with 
changes in \(\alpha_{\mathrm{count}}\).  
The estimator remains stable across all tested weights, with differences in KLD and EMD remaining small. The method is almost completely insensitive to this parameter as evidenced by very small KLD variation, very small EMD variation, and tight standard-deviation bands. This confirms that the feature-count term alone does not control the optimization trajectory. See Fig.~\ref{fig:alpha_pe} for the curves.

\begin{figure}[!htbp]
\centering
\includegraphics[width=0.45\linewidth]{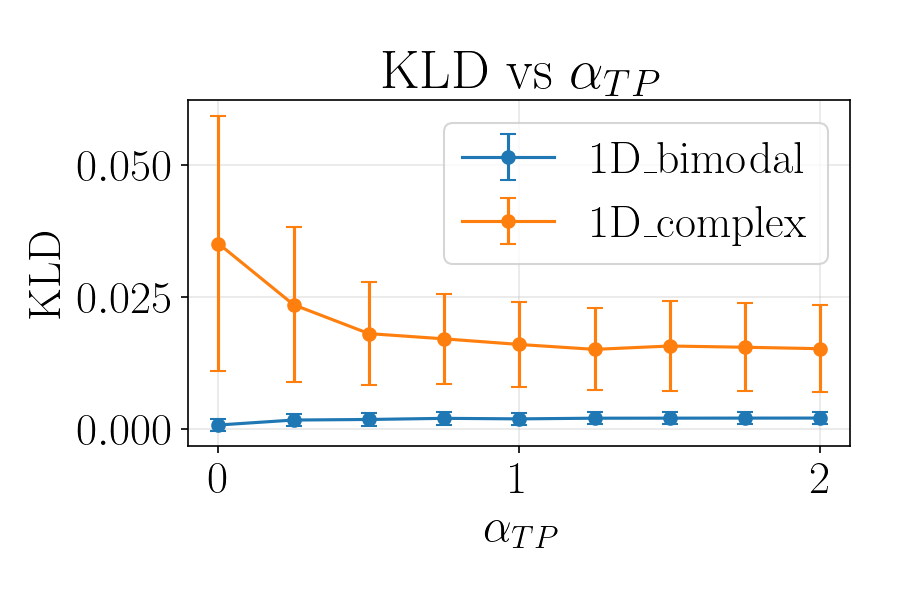}
\includegraphics[width=0.45\linewidth]{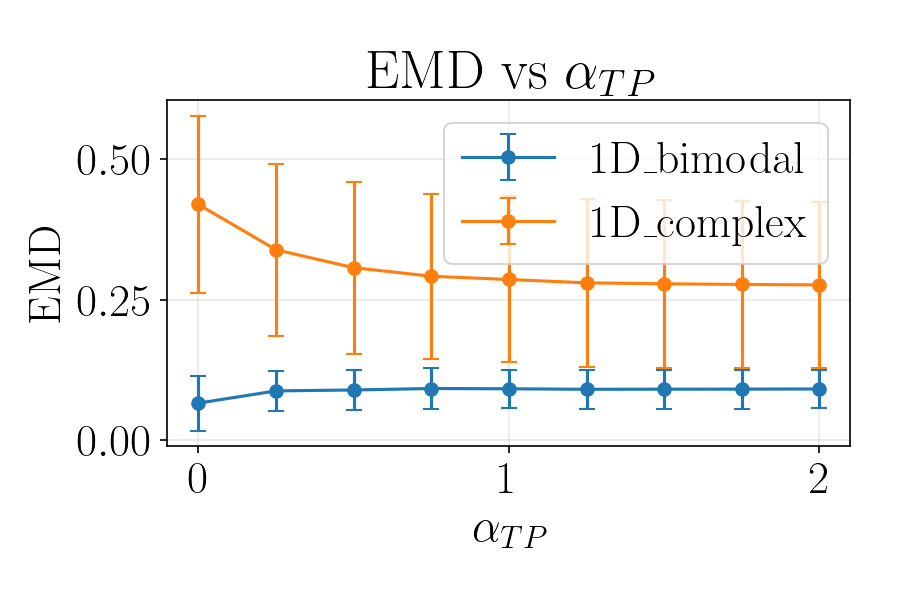}
\caption{KLD and EMD as $\alpha_\mathrm{TP}$ is varied in $[0, 2]$.}
\label{fig:alpha_tp}
\end{figure}

\paragraph{Sensitivity to \(\alpha_{\mathrm{TP}}\):}
KLD and EMD remain similarly stable as \(\alpha_{\mathrm{TP}}\) is varied.  
For the 1D Complex dataset there is a slight trend toward improved KLD values at higher weights, but the effect is modest beyond a value of 0.5. All curves remain flat with modest variance. Even large changes do not destabilize the selection. This suggests that the relative weighting of the two loss terms is not a fragile design choice. The topological objective is dominated by the structure of the persistence diagram itself rather than small re-weightings of its components.
See Fig.~\ref{fig:alpha_tp}.

\begin{figure}[!htbp]
\centering
\includegraphics[width=0.45\linewidth]{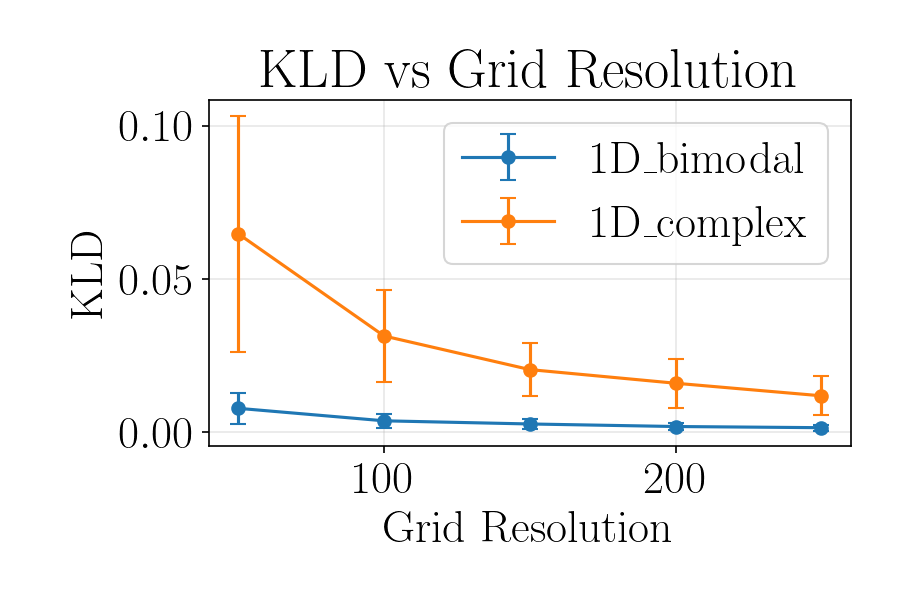}
\includegraphics[width=0.45\linewidth]{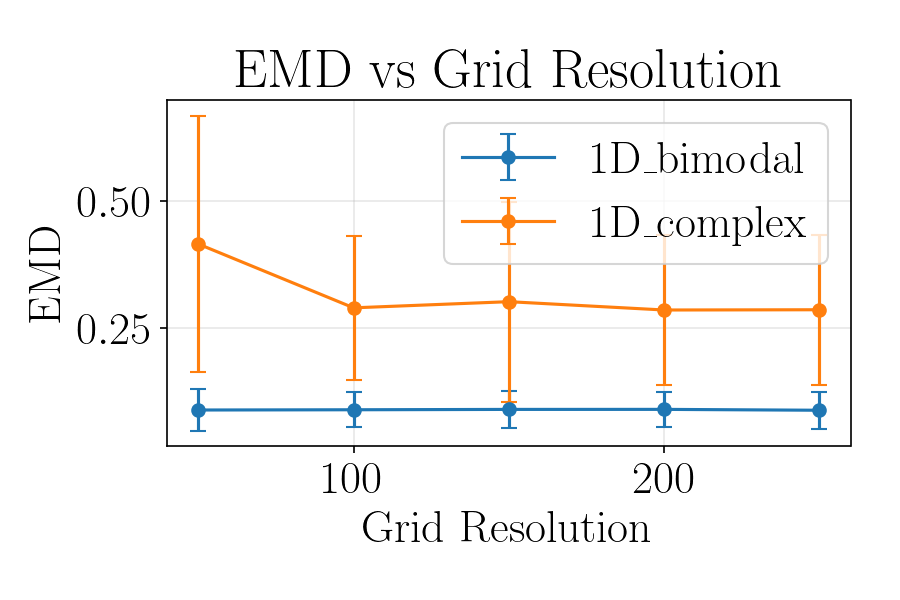}
\caption{KLD and EMD as grid size is varied in $[50, 250]$.}
\label{fig:grid}
\end{figure}

\paragraph{Sensitivity to grid resolution:}
As grid resolution increases, both KLD and EMD decrease for each dataset (particularly for the more challenging 1D complex mixture). But beyond a resolution of 100–150, improvements saturate, suggesting that very high resolutions do not lead to meaningful gains, the method is stable once a moderate grid size is reached, and memory and runtime costs can be reduced without hurting accuracy. This aligns with common sense---the loss is computed on the discretized density, so coarser grids introduce numerical imprecision that disappears once the grid is sufficiently fine.
See Fig.~\ref{fig:grid} for the curves.

Across both datasets, all hyperparameter sweeps, and 25 seeds, we find that no hyperparameter produces strong sensitivity---demonstrating that the proposed 
method is effectively \emph{tuning-free} and robust across a reasonable range of 
practical hyperparameter values.

\section{Ablation Study}
\label{sec:ablation}

To evaluate the relative contribution of each component of our optimization objective, we conducted a comprehensive ablation study, including topologically nontrivial shapes and heavy-tailed distributions. For each dataset, the bandwidth was optimized using (i) the full loss ($\alpha_{\mathrm{count}} = \alpha_{\mathrm{TP}} 1$), and (ii) single-term variants. Secondly, in principle, persistent homology produces features in all homological dimensions $H_p$, but for all our experiments we have only used $H_0$ (which corresponds to connected components)---since the number of these persistence points is typically significantly larger than points in higher dimensions. The ablation study proves that the loss is not affected by including higher dimensional homologies. We report the resulting KLD and EMD values, averaged over 100 trials.

\begin{table}[!htbp]
\centering
\scriptsize
\caption{Ablation results for all 1D and 3D datasets (mean $\pm$ std over 50 trials). 
EMD is omitted for 3D (degenerate).}
\label{tab:ablation_1d_3d}
\begin{tabular}{lcccccc}
\toprule
\textbf{Variant} 
    & \textbf{1D Bimodal KLD} 
    & \textbf{1D Bimodal EMD}
    & \textbf{1D Complex KLD}
    & \textbf{1D Complex EMD}
    & \textbf{3D Heavy-Tail KLD} \\
\midrule
full                   
    & 0.0017 $\pm$ 0.0012 
    & 0.0765 $\pm$ 0.0335
    & \textbf{0.0149} $\pm$ 0.0071 
    & 0.2721 $\pm$ 0.1085
    & 0.1663 $\pm$ 0.2164  \\
all-$H_p$                   
    & 0.0017 $\pm$ 0.0012 
    & 0.0765 $\pm$ 0.0335
    & \textbf{0.0149} $\pm$ 0.0071 
    & 0.2721 $\pm$ 0.1085
    & 0.1663 $\pm$ 0.2164  \\
no-TP                  
    & \textbf{0.0009} $\pm$ 0.0016 
    & \textbf{0.0620} $\pm$ 0.0421
    & 0.0308 $\pm$ 0.0203 
    & 0.3898 $\pm$ 0.1142
    & \textbf{0.1626} $\pm$ 0.2115  \\
no-count               
    & 0.0017 $\pm$ 0.0012 
    & 0.0757 $\pm$ 0.0336
    & 0.0154 $\pm$ 0.0072 
    & \textbf{0.2691} $\pm$ 0.1186
    & 0.1704 $\pm$ 0.2222  \\
\bottomrule
\end{tabular}
\end{table}

\begin{table}[!htbp]
\centering
\scriptsize
\caption{Ablation results for all 2D and 4D datasets (mean $\pm$ std over 50 trials).
EMD omitted for 4D (degenerate).}
\label{tab:ablation_2d_4d}
\begin{tabular}{lcccccc}
\toprule
\textbf{Variant} 
    & \textbf{2D Annulus KLD}
    & \textbf{2D Annulus EMD}
    & \textbf{2D Clusters KLD}
    & \textbf{2D Clusters EMD}
    & \textbf{4D Gaussian KLD} \\
\midrule
full                   
    & \textbf{0.0040} $\pm$ 0.0009 
    & \textbf{157.8756} $\pm$ 46.8181
    & \textbf{0.0026} $\pm$ 0.0012 
    & \textbf{62.6882} $\pm$ 22.1066
    & 0.0836 $\pm$ 0.6893 \\
all-$H_p$                   
    & \textbf{0.0040} $\pm$ 0.0009 
    & \textbf{157.8756} $\pm$ 46.8181
    & \textbf{0.0026} $\pm$ 0.0012 
    & \textbf{62.6892} $\pm$ 22.1053
    & 0.0836 $\pm$ 0.6893 \\
no-TP                  
    & 0.0055 $\pm$ 0.0002 
    & 181.4869 $\pm$ 32.3830
    & 0.0036 $\pm$ 0.0006 
    & 74.6521 $\pm$ 13.3652
    & \textbf{0.0834} $\pm$ 0.6875 \\
no-count               
    & 0.0080 $\pm$ 0.0130 
    & 160.6404 $\pm$ 43.1297
    & 0.0146 $\pm$ 0.0294 
    & 117.9810 $\pm$ 62.2989
    & 0.0838 $\pm$ 0.6909 \\
\bottomrule
\end{tabular}
\end{table}

In all low-dimensional datasets (1D and 2D), the results reveal consistent behavior: removing the TP term consistently degrades performance. In every case, the ``no-TP’’ model produces larger KLD and EMD values. By contrast, removing the betti-count penalty (“no-count’’) produces only mild changes, and usually remains close to the full loss. These results demonstrate that {the total persistence term is the dominant driver of the optimization}, while the betti-count term plays a secondary role in stabilizing the bandwidth selection. The complete loss (``full’’) therefore performs best overall, but the proximity of the ``no-count'' and ``full'' results shows that the method is not overly dependent on delicate hyperparameter balancing. In higher dimensions (3D and 4D), all variants achieve comparable KLD and EMD scores. Overall, the ablation study can be used to conclude that the {the total persistence is essential and the betti-count term is useful}. The method remains somewhat stable even when individual loss components are removed.

\end{document}